\pdfoutput=1

\documentclass[11pt]{article}

\usepackage{acl}

\usepackage{times}
\usepackage{latexsym}

\usepackage[T1]{fontenc}

\usepackage[utf8]{inputenc}

\usepackage{microtype}
\usepackage{times}
\usepackage{latexsym}

\usepackage{times}
\usepackage{url}
\usepackage{latexsym}
\usepackage{amsmath}
\usepackage{amsmath}
\usepackage{amssymb}
\usepackage{latexsym}
\usepackage{graphicx}
\usepackage{multirow}
\usepackage{breqn}
\usepackage{array,multirow}
\usepackage{tabularx}
\usepackage{pgfplotstable}
\usepackage{longtable}
\usepackage{graphicx}
\usepackage{amsmath,graphicx}
\usepackage{times}
\usepackage{latexsym}
\usepackage{times}
\usepackage{latexsym}
\usepackage{caption}
\usepackage{subcaption}
\usepackage{comment}
\usepackage{times}
\usepackage{soul}
\usepackage{booktabs}
\usepackage{url}
\usepackage[utf8]{inputenc}
\usepackage{graphicx}
\usepackage{amsmath}
\usepackage{amsmath,graphicx}
\usepackage{times}
\usepackage{latexsym}
\usepackage{times}
\usepackage{latexsym}
\usepackage{comment}
\usepackage{times}
\usepackage{soul}
\usepackage{booktabs}
\usepackage{url}
\usepackage[utf8]{inputenc}
\usepackage{graphicx}
\usepackage{amsmath}
\usepackage{amssymb}
\usepackage{array}
\usepackage{booktabs}
\usepackage{multirow}
\usepackage{multicol}
\usepackage{caption}
\usepackage{xspace}
\usepackage{amssymb}
\usepackage{booktabs}

\usepackage{multirow}
\usepackage{multicol}
\usepackage{caption}
\usepackage{xspace}
\usepackage{amsmath}

\usepackage{tabularx}

\usepackage{latexsym}
\usepackage{multirow}
\usepackage{algorithm} 
\usepackage{algpseudocode} 
\usepackage{multicol}
\usepackage{comment}
\usepackage{hyperref}
\usepackage{amsmath}

\usepackage{tabularx}

\usepackage{graphicx}
\usepackage{subcaption}
\usepackage{url}
\newcommand{\seq}{\,{=}\,}

\usepackage[linecolor=orange,size=scriptsize]{todonotes}

\newcommand{\SCC}[1]{{ \textcolor{red}{#1}}}
\newcommand{\ARR}[1]{{ \textcolor{black}{#1}}}
\usepackage{ulem}
\newcommand{\model}{\textsc{ReLiSt}\xspace}
\usepackage{lipsum} 
\definecolor{forestgreen}{rgb}{0.13, 0.55, 0.13}
\definecolor{flame}{rgb}{0.89, 0.35, 0.13}
\usepackage{enumitem}

\title{Towards Inter-character Relationship-driven Story Generation}

\author{
  Anvesh Rao Vijjini\textsuperscript{1} \hspace{.2cm}
  Faeze Brahman\textsuperscript{2,3} \hspace{.2cm} Snigdha Chaturvedi\textsuperscript{1}\\
  \textsuperscript{1}UNC Chapel Hill \\
  \textsuperscript{2}Allen Institute for Artificial Intelligence \\
\textsuperscript{3}Paul G. Allen School of Computer Science \& Engineering, University of Washington \\
  \texttt{\{anvesh,snigdha\}@cs.unc.edu} \hspace{.2cm}
  \texttt{faezeb@allenai.org} }
  
\begin{document}
\maketitle
\begin{abstract}
In this paper, we introduce the task of modeling interpersonal relationships for story generation. For addressing this task, we propose Relationships as Latent Variables for Story Generation, (\model). \model generates stories sentence by sentence and has two major components - a relationship selector and a story continuer. The relationship selector specifies a latent variable to pick the relationship to exhibit in the next sentence and the story continuer generates the next sentence while expressing the selected relationship in a coherent way. Our automatic and human evaluations demonstrate that \model is able to generate stories with relationships that are more faithful to desired relationships while maintaining the content quality. The relationship assignments to sentences during inference brings interpretability to \model.
\end{abstract}

\section{Introduction}
\label{sec:intro}
Interpersonal relationships between characters are, in many ways, the glue that holds a story together. Almost every story revolves around at least one, if not more, inter-character relationships. Despite the importance of relationships in stories~\cite{bochner1997relationships}, only few studies in NLP have explored story generation from the perspective of character relationships. Recent story generation methods typically generate stories from a prompt \cite{fan-etal-2018-hierarchical} or a planner detailing events or keywords of the story \cite{martin2018event,rashkin-etal-2020-plotmachines,goldfarb-tarrant-etal-2020-content, brahman-etal-2020-cue}. 
While these methods can generate stories based on open-ended prompts and plans they can neither encode character relationships nor can they give explicit control over the characters and their relationships.
\begin{figure}
    \centering
    \includegraphics[width=0.5\textwidth]{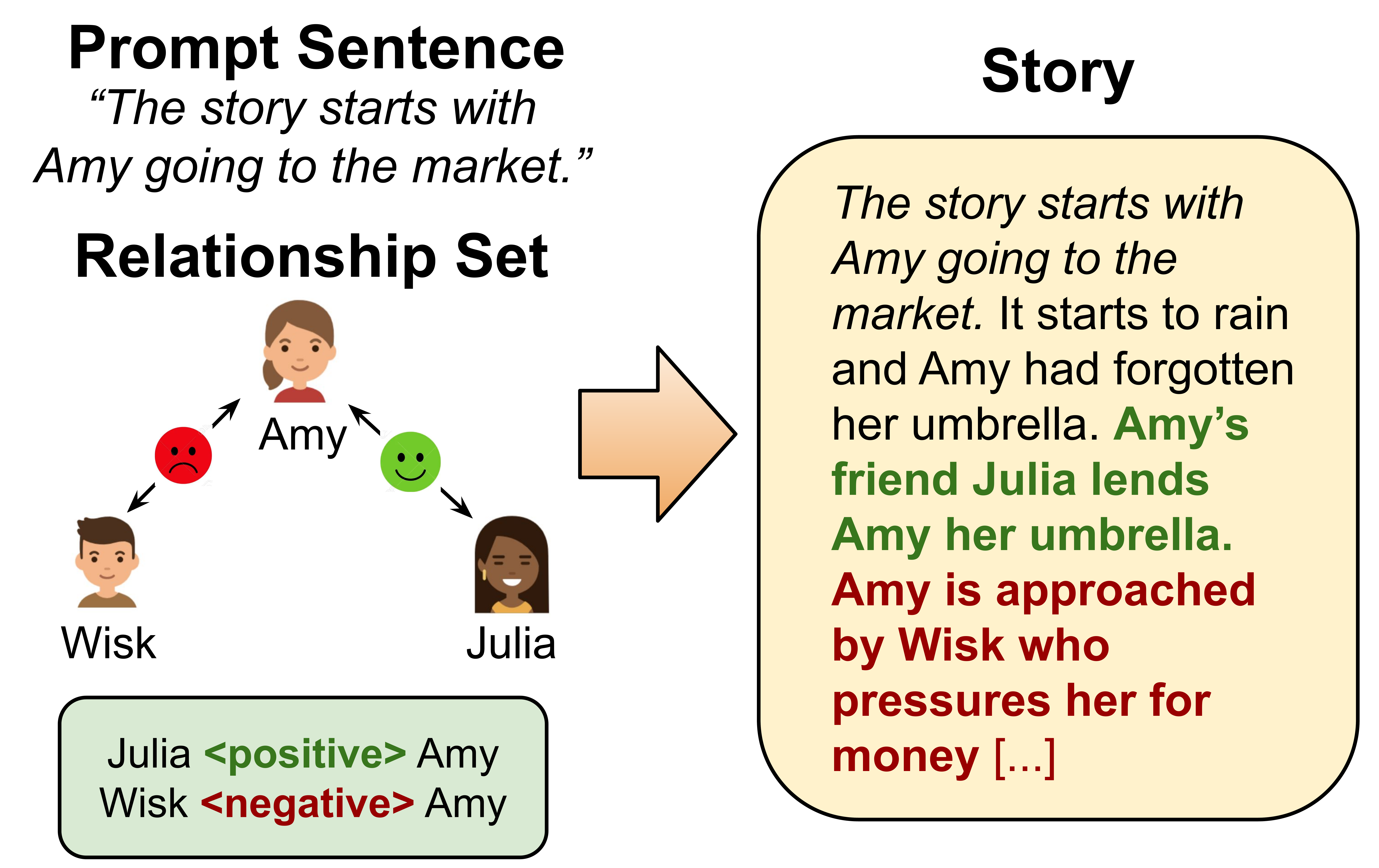}
    \caption{Example of relationship-driven story generation task: given a set of relationships and a prompt sentence, the goal is to generate a story continuing the prompt sentence and reflecting the input relationships. \textcolor{forestgreen}{\textbf{Positive}} and \textcolor{flame}{\textbf{negative}} relationships are highlighted.}
    \label{fig:example}
\end{figure}
In this paper, we introduce \textit{Relationship-driven Story Generation} where given a prompt sentence and a set of  inter-character relationships, the goal is to generate a story following the prompt sentence which exhibits the desired relationships between the characters. While relationships can be described in many ways, following previous works~\cite{chaturvedi2016modeling, srivastava2016inferring, si-etal-2021-telling}, we represent relationships using \textit{relationship polarity}.  Specifically, we summarize the overall interaction between pairs of characters as being \textit{positive, neutral} or \textit{negative}. Figure \ref{fig:example} illustrates the task. 

Apart from the challenges of story generation in general, there are several challenges unique to the proposed task. The first challenge is of \textit{relationship selection}. In a typical story, some sentences describe interpersonal relationships while others do not. For example, in the narrative shown in figure \ref{fig:example}, the first sentence after prompt sentence does not exhibit any relationship unlike the second and third sentences which express different relationships.  Hence, the story generation model needs to decide \textbf{when} to exhibit \textbf{which} relationship based on the context. The second challenge is of \textit{story continuation}. The way characters behave toward each other defines their relationships. Therefore, the model needs to generate events that naturally reflect the desired relationships while maintaining the overall coherence of the narrative. Both these challenges require the model to capture long-range dependencies across multiple sentences and characters.
We approach these challenges by a modeling framework which treats relationships as latent variables. Our proposed model, \model, generates stories sentence by sentence where each sentence is associated with a latent variable that encodes which, if any, relationship is exhibited in the sentence. \model has two components: the \textit{relationship selector} and the \textit{story continuer}. The relationship selector explicitly handles the aforementioned \textbf{when} and \textbf{which}  challenges. Specifically, before generating any sentence, the relationship selector selects which relationship (or no relationship) to be used for conditioning the next sentence's generation. The story continuer then generates a sentence which naturally reflects the selected relationship while ensuring logical continuation of the narrative. These components work together to produce a naturally coherent story and are trained jointly. 

We define two automatic and reference-less metrics to measure relationship faithfulness, i.e. the models' ability to generate stories that are faithful to the input relationships. We assess the content quality and relationship faithfulness through automatic and human evaluations. Our results show that \model can express the desired relationships while maintaining fluency and coherence in the generated stories. We provide additional analyses where we leveraged our latent variable based design to get insight into the generation process. We summarize our contributions as follows:
\begin{itemize}[noitemsep,topsep=0.3pt]
\item We present the first study on relationship-driven story generation using interpersonal relationships as controllable parameters.
\item We propose \model, made up of two components - \textit{relationship selector} and \textit{story continuer} which are trained jointly.
\item We conduct automatic and human evaluations to empirically demonstrate that \model can not only reflect desired relationships but also generate fluent stories. 
\item Our analysis shows how \model can offer transparency to the generation process.
\end{itemize} 
\section{Related Work}
Story generation was first approached with symbolic planning \cite{perez2001mexica,porteous2009controlling,riedl2010narrative}. Since then, neural networks have gained more interest for story generation. Several of these methods generate stories using coarse-grained prompts or outlines~\cite{fan-etal-2018-hierarchical,yao2019plan, brahman-etal-2020-cue,rashkin-etal-2020-plotmachines,guan-etal-2020-knowledge} or event-based plans~\cite{martin2018event,fan-etal-2019-strategies, goldfarb-tarrant-etal-2020-content}

Besides the development of plan through keywords or events, there are other elements that contribute to a good story. Characters~\cite{bamman:2013,bamman:2014,brahman-etal-2021-characters-tell,zhang2019generating,azab-etal-2019-representing}, their sentiment trajectory~\cite{chaturvedi-etal-2017-story} and relationships with others~\cite{kim-klinger-2019-frowning,iyyer-etal-2016-feuding, chaturvedi2017unsupervised} have been shown to be useful for story understanding, in general. For example, \newcite{si-etal-2021-telling} proposed using interpersonal relationships for predicting the best story continuation in first person dialogue-based stories. Characters have also been shown to be useful for language generation such as in dialogue systems for generating responses conditioned on character's persona \cite{majumder-etal-2020-like,majumder-etal-2021-unsupervised,li-etal-2016-persona,oraby-etal-2018-controlling}. Nevertheless, only a few works have modeled characters for story generation. Existing character-centric storytelling systems have conditioned generation on automatically learnt character embeddings \cite{liu2020character} or persona~\cite{chandu-etal-2019-way,zhang-etal-2022-persona}. We instead model characters through their interpersonal relationships.

Our work is also related to the problem of text generation conditioned on sentiment or emotions. ~\citet{peng-etal-2018-towards} and \citet{luo-etal-2019-learning} proposed controlling the sentiment for story ending generation. \citet{scaffold:20} proposed incorporating sentiment for the story in-filling task. \newcite{brahman-chaturvedi-2020-modeling} generate stories that adhere to desired emotional arcs for the protagonist. 
In story generation, \citet{jhamtani-berg-kirkpatrick-2020-narrative} use latent ``anchor words'' in each sentence as a plan that generates the story. \citet{xie-etal-2021-exploring} employ variational autoencoders to generate stories with informative latent variables for more diverse and coherent story generation. \model provides control over relationships in story generation by considering relationship expressed in a sentence as a latent variable.


\section{Relationship-driven Story Generation}
In this section, we first present a formal description of our task (Sec. \ref{sec:problem}) followed by model design of \model and its training procedure (Sec. \ref{sec:latent_var}).

\subsection{Problem Statement}
\label{sec:problem}
In the relationship-driven story generation task, given a prompt sentence and a set of interpersonal relationships, the goal is to generate a story following the prompt sentence which reflects the desired relationships. More formally, let $x_0$ be the prompt sentence, and $\mathfrak{R}=\{r^{j}\}_{j=1}^K$ be a set of interpersonal relationships. Each $r^{j}$ is a triple in the form of \texttt{Char$_1$ <$P$> Char$_2$}, where \texttt{<$P$>} $\in$ $\{positive, neutral, negative\}$ indicates the polarity of the relationship (``relationship polarity'') between \texttt{Char$_{1}$} and \texttt{Char$_{2}$}. 
Given $x_{0}$ and $\mathfrak{R}$ as inputs, the goal is to generate a story $S = \{x_{1}...x_{T}\}$ where $x_{i}$ is the $i^{th}$ sentence of the generated story. The generated story should include the specified characters and reflect their relationships in $\mathfrak{R}$ while being narratively coherent and manifesting a natural progression. 

\subsection{Relationships as Latent Variables for Story Generation}

\label{sec:latent_var}
We propose \textbf{Re}lationships as \textbf{L}atent Var\textbf{i}ables for \textbf{St}ory Generation 
(\model). \model generates stories sentence by sentence and views the relationship exhibited in a sentence as a latent variable. \model is composed of two components, namely \textit{relationship selector} and \textit{story continuer}. 

Given the story generated so far (``context''), the \textit{relationship selector} decides whether, and if so, which relationship to express in the next sentence. The relationship expressed in each sentence, $x_i$, is modeled as a discrete latent variable, $z_i$.  
 Since in a typical story, not all sentences express inter-character relationships, we introduce an additional value for $z_i$ describing  a ``null'' relationship, $\emptyset$. Hence, $z_i \in {R}$ where ${R} = \{r^{j}\}_{j=0}^K$, $r^{0}=\emptyset$  and in the generative process, $z_i$ is sampled from relationship selector before generating the next sentence.
Formally, the relationship selector models $p(z_i|C_{<i},R)$ which is the probability distribution of discrete random latent variable $z_i$, over the interpersonal relationship set ${R}$. $C_{<i}=\{x_0, x_1, ...x_{i-1}\}$ denotes the context or story so far. We parameterize this component using a classifier with parameters $\phi$. 
\begin{figure}[!t]
    \centering
    \includegraphics[width=0.45\textwidth]{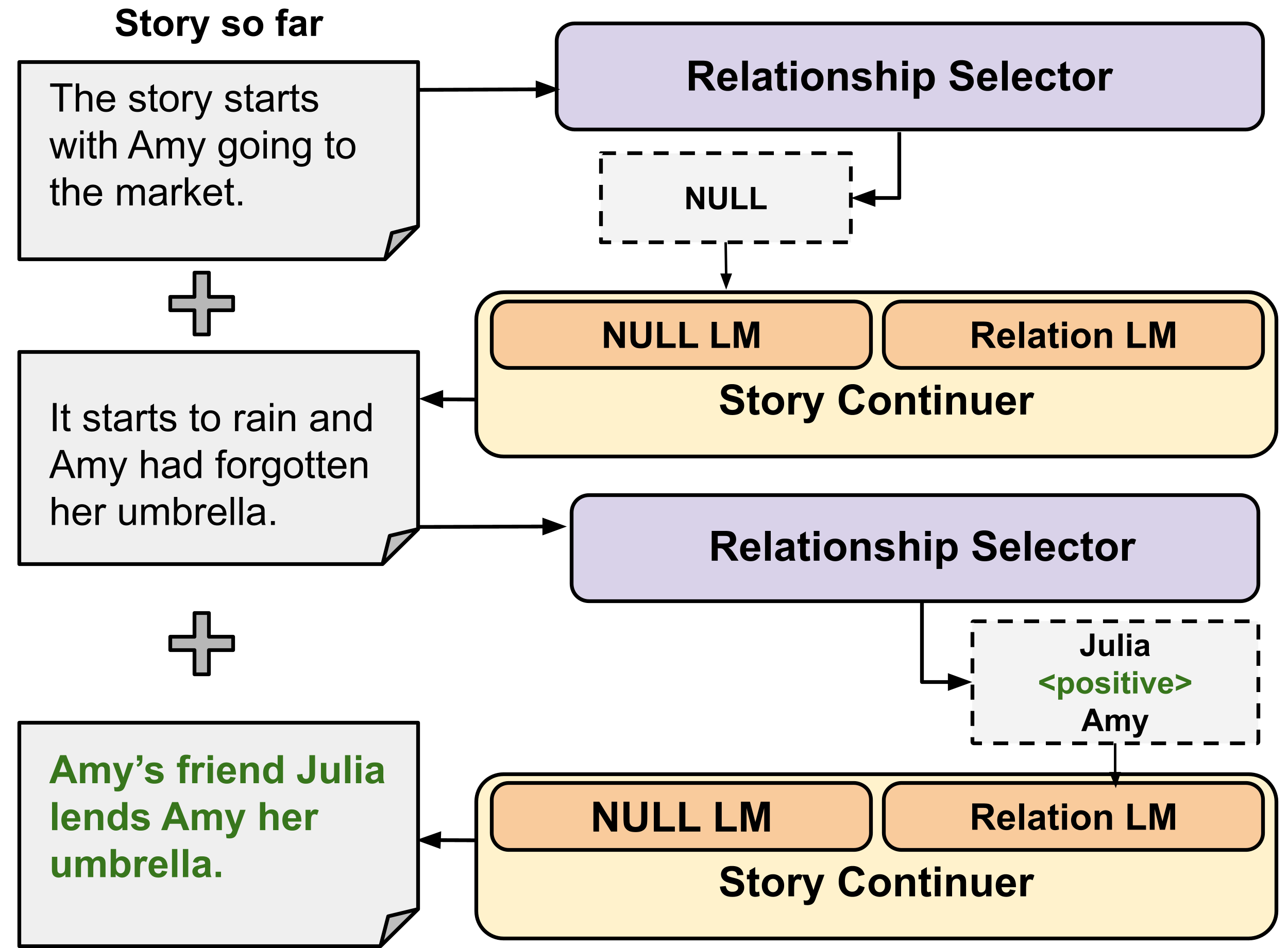}
    \caption{Proposed model \model illustrated. \model has two components,  the relationship selector and the story continuer,  which jointly generate the story.}
    \label{fig:model}
\end{figure}
Conditioning on the relationship selected by the \textit{relationship selector}, the \textit{story continuer} generates the next sentence of the story while maintaining coherence to the context. Formally, the story continuer models $p(x_{i}|z_i,C_{<i},R)$ which is used to sample the next sentence, $x_{i}$\, given a relationship, $z_{i}$, the context $C_{<i}$, and the set of interpersonal relationships, $R$. 
To distinguish between the case when the sentence exhibits a relationship ($z_i$ $\in$ $\mathfrak{R}$) and when it does not ($z_i = \emptyset$), we use two distinct language models-- ``Relationship LM'' and ``Null LM'', respectively. Collectively, the parameters of this component is represented as $\theta$. Figure ~\ref{fig:model} shows our model architecture.
\newline
\noindent \textbf{Training \model.}\space\space
In relationship-driven story generation, we aim to model $p(S|x_{0},R)$. Using the chain rule, the likelihood can be written as:
\begin{equation}
     P(S|x_{0},R) = \prod_{i=1}^{T} p(x_{i}|C_{<i}, R)
\end{equation}
\noindent Using the discrete latent relationship variable $z_i$, the likelihood can be rewritten as:
\begin{equation}
     \prod_{i=1}^{T} \sum_{j=0}^{K} p_{\theta}(x_{i}|z_i \seq r^{j}, C_{<i}, R)p_{\phi}(z_i\seq r^{j}|C_{<i}, R)
\end{equation} Here, $\phi$ and $\theta$ represent the parameters of the \textit{relationship selector} and the \textit{story continuer} respectively. The two components are trained jointly using Expectation Maximization \cite{DEMP1977}.
\noindent

In the E-step, we estimate the expected posterior for the latent variables $p(z_i|x_i, C_{<i}, R)$ (via the Bayes Rule) as:\begin{equation}
\label{eq3}
     p(z_i|x_{i},C_{<i},R) \propto p_{\theta}(x_{i}|z_i,C_{<i},R)p_{\phi}(z_i|C_{<i},R)
\end{equation} \noindent where, $x_{i}$ is a story sentence and $z \in {R}$. The expectation for latent variable assignments $z_{i}$ is estimated using parameter values ($\phi$ and $\theta$) of the previous iteration. Using $p(z_{i}|x_{i},C_{<i},R)$, we sample new updated assignments of $z_{i}$ for each sentence $x_i$.

In the M-step, given the new latent variable assignments, we maximize the following log likelihoods to update the parameters $\phi$ and $\theta$. Specifically, \ARR{we train \textit{relationship selector} by optimizing:} 
\begin{equation}
     L(\phi) =   \sum_{i=1}^T log \, p_{\phi}(z_{i}|C_{<i},R) 
\end{equation}
We train \textit{story continuer} by optimizing:
\begin{equation}
L(\theta) =  \sum_{i=1}^T log \, p_{\theta}(x_{i}|z_{i},C_{<i},R)
\end{equation}
\vspace{10pt}\newline
\noindent \textbf{Implementation Details.} \space\space
\model has three neural networks used in the relationship selector and the story continuer (``Relationship LM'' and ``Null LM'').\ARR{ Since there are varying number of relationships, the relationship selection task cannot be addressed using a simple classifier. We instead train a BERT model \cite{devlin-etal-2019-bert} to receive input as \texttt{${R}$  [SEP] $C$} and output the start and end pointers indicating the start and end tokens for the next relationship from the tokens in ${R}$. Thus, the model is able to choose from the relationship set or \textit{null} by pointing to either of them. 
This setting is similar to how BERT was proposed for Question Answering.} 
For the story continuer, we use decoder-only Transformers initialized with GPT-2-medium \cite{radford2019language}. Specifically, the ``Relationship LM'' is trained to receive inputs as \texttt{$y\mathfrak{R}$ $<$@$> C_{<i} <$@$> z_i <$\$$> $} \ARR{where $z_i$ $\in$ $\mathfrak{R}$} and outputs the next sentence, $x_i$. The ``Null LM'' is trained to receive inputs as \texttt{$\mathfrak{R} <$\$$> C_{<i}$} and outputs the next sentence, $x_i$. 
We initialize each of the three neural networks by training them on stories with sentences automatically annotated with relationships. For obtaining these annotations, we assign a relationship from ${R}$ to each sentence in the story using the pipeline described in Sec. \ref{sec:dataset}. The ``Relationship LM'' is trained on sentences $x_i$ which are annotated with some relationship $z_i$ $\in$ $\mathfrak{R}$.  The ``Null LM'' is trained on all sentences allowing it to learn fluency and coherence. 
In practice, we delay updating $\theta$ (story continuer parameters) until $warmup$ iterations of the E and M steps.\footnote{We found $warmup =1$ and total EM cycles $E=3$ to be helpful in improving the generation quality of LMs} This is because a noisy relationship selector model ($p_{\phi}(z_i|C_{<i},R$)) influences story continuer ($p_{\theta}(x_{i}|z_i,C_{<i},R$) from Equation \ref{eq3}. This implies that sentences which do not correspond to any of the interpersonal relationships might be assigned to one of them (noisy latent variable assignments). Maximizing the objective of ``Relationship LM'' with the noisy latent variable assignments results in poor relationship faithfulness. 


\section{Experimental Setup} 
In this section we explain the dataset used for training and initialization of \model (Sec. \ref{sec:dataset}) followed by baselines (Sec. \ref{sec:baselines}) and evaluation measures (Sec. \ref{sec:eval}).
\subsection{Dataset and Annotation Pipeline}
\label{sec:dataset}
\label{subs:data}
We use the CMU Movie Summary Corpus \cite{bamman-etal-2013-learning} for our experiments. It contains $42,306$ movie summaries as stories. Each story has on average $375$ words. 
For our experiments, we need the stories labeled with interpersonal relationships. For this, we automatically annotate stories in the CMU Movie corpus with interpersonal relationships. We refer to this automatically annotated corpus as the \textit{silver labelled} dataset and it is created using the following pipeline.
First, we process the stories with the BookNLP~\cite{bamman-etal-2014-bayesian}\footnote{\url{https://github.com/dbamman/book-nlp}} toolkit to identify dependency parse labels and character mentions. Second, we identify sentences with two character mentions with the constraint that one of the character is the subject and the other is the object of the main verb. This constraint helps in capturing the interpersonal relationship and not the overall sentiment of a sentence. For example, ``\textit{John and Beth lost all their money.}'' has an overall negative sentiment but does not indicate a negative relationship between John and Beth. Third, we concatenate all sentences containing mentions of the same pair of characters as a global representation of their interactions. Finally, we obtain the overall relationship polarity for the character pair using the sentiment of the combined sentences. For this, we use the Sentiment Intensity Analyzer toolkit.\footnote{\url{https://www.nltk.org/howto/sentiment.html}} The toolkit returns intensity scores for ``positive'', ``neutral'' and ``negative''.

Using this pipeline, we annotate each story with the polarities of interpersonal relationships between all pairs of character mentions. We discard stories for which we are unable to identify any interpersonal relationships. Our final \textit{silver labelled} data contains $16,886$  stories ($14,712$ train and $2,174$ test) with an average of $2$ relationships per story. The dataset contain $31,488$ relationships in total. The distribution of polarities is ``positive'' $36.21\%$, ``neutral'' $18.78\%$ and ``negative'' $45.01\%$.\ARR{Appendix \ref{app:quality-data} presents details about the quality assessment of this automatically annotated data} 




The \textit{silver labelled} dataset can also be extended to obtain sentence-level relationship annotations. For this, sentences with two character mentions selected in the second step of the pipeline are annotated with their corresponding overall relationship polarity identified in the last step of the pipeline. 
Sentences not selected in the second step are annotated with $\emptyset$. This data is used to initialize the story continuer and the relationship selector in \model. 

\subsection{Baselines}
\label{sec:baselines}
We compare \model against the following story generation models. All baselines use the \textit{silver labelled} dataset. Furthermore, the initialization process of the Language Models of \model also uses sentence level relationship annotation. Hence, for fair comparison we have also included a baseline (GPT-2 Planned) using this information.

\noindent\textbf{Fusion.} \space \citet{fan-etal-2018-hierarchical} use a seq2seq architecture to generate stories conditioned on natural language prompts. 
To use this method, we concatenate natural language descriptions for the relationships and the first sentence of the story into a single prompt. 
 The model is trained to continue the story conditioned on this prompt.

\noindent\textbf{Plan and Write (PW).} \space\space
 \newcite{yao2019plan} use two seq2seq architectures to first generate a plan from the title, and then generate the story from the plan. We adapt their model by concatenating the set of relationships and the prompt sentence as a plan and using it to generate the rest of the story. 

\noindent\textbf{BART FT.}\space\space
BART~\cite{lewis-etal-2020-bart} has shown success in story generation~\cite{goldfarb-tarrant-etal-2020-content}. We finetune BART-large for our task using the concatenation of relationship set and the prompt sentence as input and rest of the story as output. 

\noindent\textbf{GPT-2 FT.} \space\space
We additionally fine-tune a standard left-to-right LM, namely GPT-2-medium \cite{radford2019language}, to generate the story conditioned on the relationship set and the first sentence as prompt. 
\noindent\textbf{GPT-2 Planned.} \space\space
\ARR{We also compare \model against a planner based GPT-2 model. This baseline includes two components: 1. a planner which generates a sequence of relationships corresponding to each sentence in the potential final story, given the input relationship set $R$ (including $\emptyset$) and the first sentence. 2. a generator which generates the full story given the generated sequence of relationships. Both components are initialized with GPT-2-medium and finetuned using the sentence-level annotations obtained in Section \ref{sec:dataset}.}

\noindent\textbf{\model-$0$.} \space\space
This is \model after initialization. The individual components are trained with the \textit{silver labelled} data but there is no joint training after that.

\subsection{Evaluation Measures}
\label{sec:eval}
We use automatic and human evaluation to assess the efficacy of our model. We evaluate how faithful are the generated stories w.r.t the input relationships and the content quality of the generated stories. 

\subsubsection{Automatic Evaluations}

\noindent\textbf{Relationship Faithfulness Metrics}:
Relationship faithfulness describes measures how faithfully the model generates stories that express the desired relationships in the input. 
To evaluate relationship faithfulness, we propose two reference-less automatic metrics:

\vspace{10pt}
\noindent\textbf{1. Relationship Identification (RI):}
Ideally, stories generated from a relationship-faithful model should reflect the desired relationships. Therefore, in ``Relationship Identification'' metric, we compare relationships exhibited in the model-generated stories to the input relationships. For this, we run the annotation pipeline (Sec. \ref{sec:dataset}) on the generated stories to identify relationships. We then compare the identified relationships to the corresponding input relationship sets and compute the following: 
\begin{itemize}[noitemsep,topsep=0.3pt]
    \item \%Exact: The percentage of relationships identified in the generated stories that exactly match with the input relationships.
    \item \%Unspec: The percentage of relationships identified in the generated stories \ARR{that contain character pairs not specified in the input relationship set. Hence, these relationships do not reflect the desired ones}.
    \item \%Incorrect: The percentage of relationships identified in the generated stories that contain correct character pairs but incorrect polarity of their relationships. 
    \item Average Relationships (AvgRel): Average number of identified relationships in the generated stories
\end{itemize}

\begin{table*}[!t]
\setlength\tabcolsep{4.2pt}
\centering
\footnotesize
\begin{tabular}{lccccc}
\toprule
\textbf{Models} & \textbf{\%Exact} ($\uparrow$) & \ARR{\textbf{\%Unspec}} ($\updownarrow$) & \textbf{\%Incorrect} ($\downarrow$) & \textbf{AvgRel} ($\uparrow$) & \textbf{P-CLS} ($\uparrow$)\\
\hline
Fusion \cite{fan-etal-2018-hierarchical} & 28.32 &16.33  & 55.37& 0.32 &42.37 \\
PW \cite{yao2019plan} & 24.60 & 17.21& 58.19 & 0.78 &42.16 \\
BART FT \cite{lewis-etal-2020-bart} & 28.48 & 27.72& 43.80& 1.51 &47.31 \\
GPT-2 FT \cite{radford2019language} &  33.31& 35.95 & 30.75& 1.48 & 46.48 \\ 
GPT-2 Planned  & 32.62 & 38.94 & 28.44& 1.44 & 46.25\\\hline
\model-$0$ & \textbf{43.61} & 29.30 & 27.09 & 1.09 &45.13 \\ 
\model &42.85 & 31.07 & \textbf{26.08} & \textbf{1.57} &  \textbf{50.32}  \\
\hline
\end{tabular}\\
\begin{tabular}{lcccccc}
\hline
\textbf{Models} & \textbf{BLEU} ($\uparrow$) & \textbf{R-1} ($\uparrow$) &\textbf{R-L} ($\uparrow$) & \textbf{Dist-1} ($\uparrow$) & \textbf{Dist-2} ($\uparrow$) & \textbf{Dist-3} ($\uparrow$)\\
\hline
Fusion \cite{fan-etal-2018-hierarchical} &  20.84 &28.40 & 26.71& 80.32& 93.38&99.01 \\
PW \cite{yao2019plan} & 21.73& 26.68& 25.36& 81.93 & 94.57&99.38\\
BART FT \cite{lewis-etal-2020-bart} & 28.93 &\textbf{30.22} & \textbf{28.24}& \textbf{83.49} & 93.87& 99.63 \\
GPT-2 FT \cite{radford2019language} & \textbf{29.27} & 27.76 & 26.38& 82.51 & 94.33 & \textbf{99.69}\\
GPT-2 Planned& 22.51& 26.14& 25.36 & 82.13 & 93.13 & 99.25\\\hline
\model-$0$ & 22.45 & 26.22 & 25.15 & 81.87 & 92.75 & 99.06 \\ 
\model & 28.77 & 27.84 &26.44& 83.38 & \textbf{95.21} & 99.54 \\
\hline
\end{tabular}
\caption{\label{tab:auto_results2} Automatic evaluation results for relationship faithfulness (top) and content quality (bottom). In relationship faithfulness, \model and \model-$0$ outperform all baselines. Low AvgRel of \model-$0$ indicates ``Relationship collapse'' whereas \model generates higher number of relationships with high \%Exact. In content quality,  \model is comparable to baselines. }
\end{table*} 

\noindent\textbf{2. Polarity Classification (P-CLS)}:
If the stories generated by a model accurately reflect the desired relationships, then they can be used for the `inverse' task of identifying relationship polarities from the text of the stories. Such stories can be used to train a polarity classifier that takes as input a generated story and a character pair, and outputs the relationship polarity label between the characters. \ARR{This polarity classifier can then be evaluated on a subset of 105 stories~\cite{srivastava2016inferring} that are manually annotated with relationships. The classifier accuracy will indicate the goodness of its training data-- how well the generated stories reflected the desired relationships. We define P-CLS as the accuracy of this classifier.}

For training the polarity classifier, we finetune BERT-base~\cite{devlin-etal-2019-bert} on stories generated by the model. We provide more training details in Appendix~\ref{app:details}.

\noindent\textbf{Content Quality Metrics}: 
For automatic assessment of content quality, we use the standard $n$-gram overlap based metrics such as BLEU~\cite{papineni-etal-2002-bleu}, and ROUGE scores~\cite{lin-2004-rouge}.
For evaluating diversity of generation, we use Distinct-$n$ ($n=1,2,3$) to measure the percentage of unique $n$-grams~\cite{li-etal-2016-diversity}. Higher score implies higher lexical diversity. 

\begin{table}
\centering
\footnotesize
\setlength\tabcolsep{6pt}
\renewcommand{\arraystretch}{1.1}
\begin{tabular}{lc}
\hline
\multirow{2}{*}{\textbf{Criteria}} & \textbf{Preference}\\
 &Better/Worse/Tie (\%)\\
\hline
Relationship Faithfulness & $50.00^\ast\:/\:30.00\:/\:20.00$\\
Content Quality & $41.67\:/\:43.33\:/\:15.00$ \\
Overall & $48.33^{**}\:/\:33.33\:/\:18.33$ \\
\hline
\end{tabular}
\caption{\label{tab:human_results} Human evaluation results for \model vs. GPT-2 FT. \model better expresses desired relationships and is overall preferred. $^*$ and $^{**}$ denote the difference is significant with $p < 0.01$ and $p< 0.07$ via t-test.}
\end{table}

\subsubsection{Human Evaluations}
We also conduct human evaluation of the generated stories using Amazon Mechanical Turk. We show workers a pair of stories generated from two different models using the same prompt sentence and relationship set as input. We then ask the workers to compare pair of stories based on three criteria: (1) content quality, (2) relationship faithfulness, and (3) overall. For each criteria, the workers have to pick the better story of the two or indicate that they are of equal quality. For the overall criteria, the workers are explicitly asked to consider both relationship faithfulness and content quality. Appendix~\ref{app:human-eval} contains detailed instructions. 

\begin{table*}[!t]
\setlength\tabcolsep{4.2pt}
\centering
\footnotesize
\begin{tabular}{lccccc}
\toprule
\textbf{Models} & \textbf{\%Exact} ($\uparrow$) & \ARR{\textbf{\%Unspec}} ($\updownarrow$) & \textbf{\%Incorrect} ($\downarrow$) & \textbf{AvgRel} ($\uparrow$) & \textbf{P-CLS} ($\uparrow$)\\
\hline
\model-$SLM$ & 31.48 & 28.61 & 39.91 & 1.36 & 42.27  \\
\model-$RandSelect$ & 44.58 & 23.51 & 31.91 & 1.44 & 44.61  \\ \hline
\model &42.85 & 31.07 & \textbf{26.08} & \textbf{1.57} &  \textbf{50.32}  \\
\hline
\end{tabular}\\
\begin{tabular}{lcccccc}
\hline
\textbf{Models} & \textbf{BLEU} ($\uparrow$) & \textbf{R-1} ($\uparrow$) &\textbf{R-L} ($\uparrow$) & \textbf{Dist-1} ($\uparrow$) & \textbf{Dist-2} ($\uparrow$) & \textbf{Dist-3} ($\uparrow$)\\
\hline
\model-$SLM$ & 22.51 & 27.21 & 24.74 & 81.64 & 93.86 & 99.25 \\
\model-$RandSelect$ & 21.05 & 26.85 & 25.51 & 80.69 & 92.28 & 99.25 \\ \hline
\model & 28.77 & 27.84 &26.44& 83.38 & \textbf{95.21} & 99.54 \\
\hline
\end{tabular}
\caption{\ARR{\label{tab:ablate2} Ablation studies results for relationship faithfulness (top) and content quality (bottom).}}
\end{table*}

\section{Experimental Results}
Here, we first report the automatic and  human evaluation results, followed by ablations and analyses of \model. 

\subsection{Relationship Faithfulness}

Table \ref{tab:auto_results2} shows the results of automatic evaluation for relationship faithfulness. We observe that \model and \model-$0$ outperform other baselines by generating stories with higher \%Exact and lower \%Incorrect scores. 
BART-FT and GPT-2 FT  tend to introduce \ARR{unspecified} relationships (high \%Unspec) compared to other models but this is not necessarily an indicator of poor performance. They also generate higher \%Incorrect relationships with comparable AvgRel showing that they have poor control over polarity of desired relationships as compared to \model.

By comparing GPT-2 FT and \model, we can infer that introduction of latent variables increases relationship faithfulness since that is the major difference between these models. Since \model LMs are parameterized by GPT-2 architecture and initialization, any performance due to Language Modeling in Relist will also be reflected in GPT2. However, we see significant improvement between the two, leading us to credit latent variables.  




\ARR{Comparing \model and \model-$0$ shows the effectiveness of iterative training after intialization with \textit{silver labelled} data.} Although \model-$0$ has high \%Exact and low \%Incorrect values, it generates significantly less number of relationships (low AvgRel). \model-$0$ suffers from what we call ``Relationship collapse'' problem where the relationship selector repeatedly picks the same relationship from the input. This results in generating fewer relationships in the story. On the other hand, \model achieves good \%Exact and \%Incorrect scores while having high AvgRel. 
\ARR{The GPT-2 Planned is not sufficient for relationship faithfulness. This is because unlike \model 's relationship selector, the first LM generates relationships without having access to prior generated sentences.
}

Table \ref{tab:auto_results2} also shows that \model achieves the highest P-CLS score. The polarity classification task  is challenging because the training supervision for the  polarity classifier is obtained from automatically annotated data but it is tested on human annotated data. As a result, most baselines are near random performance of $42.15$\footnote{Note that random performance $\approx 42$ because of imbalanced polarity distribution as explained in Section \ref{sec:dataset}}, whereas \model achieves a higher P-CLS score of $50.32$. 
Overall, the results for relationship faithfulness demonstrate \model's superiority at generating stories expressing the desired relationships. 

\begin{table*}[h]
\scriptsize
\renewcommand{\arraystretch}{1.2}
\begin{tabularx}{\textwidth}{lX}
    \hline
\textbf{$R$} & Jonny \textcolor{forestgreen}{<positive>} Curtis\\
\textbf{Prompt} &  The film centers around Jonny Barkley, an ambitious, image - conscious businesswoman working for over-demanding boss Mary.  \\
     \hline
    \textbf{GPT-2 FT} &  The film centers around Jonny Barkley, an ambitious, image - conscious businesswoman working for over-demanding boss Mary.  Jonny's life is in turmoil when she is fired from her job, and she is forced to move to New York City, where she is working as a waitress. Jonny has a crush on her, but she doesn't know that she is pregnant. When she learns that Jonny is pregnant, she decides to take matters into her own hands. She goes to her boss's office to meet with him, and he tells her that he wants her to have an abortion. When Jonny refuses, the boss tells her to go to the [...]\\ \hline
\textbf{\model} & The film centers around Jonny Barkley, an ambitious, image - conscious businesswoman working for over-demanding boss Mary. One day, Jonny is approached by Curtis, who wants to hire Jonny as his personal assistant. Curtis tells Jonny that he's been offered a job in New York City, but Jonny declines Curtis'offer. Jonny and Curtis go to a bar, where Jonny meets Curtis's ex-boyfriend. Jonny tells Curtis that he loves [...]
 \\
    \hline
     \hline
\end{tabularx}
  \caption{In the example, unlike GPT-2 FT which makes no mention of ``Curtis'' and a \ARR{unspecified} relationship with ``Jonny'' and ``Mary'', \model faithfully described the input relationship ``Jonny <positive> Curtis''. }
  \label{tab:examples}
\end{table*}

\subsection{Content Quality}

Table \ref{tab:auto_results2} shows the results of automatic evaluation for content quality. \ARR{GPT-2 Planned}, GPT-2 FT, BART FT, and \model are comparable in content quality. 
\model-$0$ suffers from coherence significantly. As mentioned before, it suffers from ``Relationship collapse'' where the model chooses the same relationship repeatedly. As a result, the story continuer faces a challenge in maintaining coherence while expressing the relationship in every sentence. \model's generated stories do not reflect this problem.
The results for content quality and relationship faithfulness show that \model solves the task of relationship-driven story generation while maintaining fluency and coherence. 

\subsection{Human Evaluation}

Since automatic metrics do not evaluate all aspects of open-ended NLG~\cite{novikova-etal-2017-need}, 
We also conduct human evaluation for better evaluation of open-ended NLG. For this, we randomly sample 50 instances from test set described in Sec. \ref{sec:dataset}, and use their relationship sets and prompt sentences to generate stories from \model and GPT-2 FT which is among the strongest baseline according to automatic evaluation. Each story pair was evaluated 
for relationship faithfulness, content quality and overall preference. The annotators read 100 stories in total. Human evaluation for this problem is particularly time consuming and laborious. It requires the annotators to thoroughly read two long stories, then compare them for relationship faithfulness based on the provided relationship set and content quality.

Table \ref{tab:human_results} shows the results for each criteria. We see that \model is better than GPT-2 FT at relationship faithfulness. These results also validate the automatic metrics (RI and P-CLS) proposed for relationship faithfulness. 
While \ARR{the difference between the two models in terms of content quality is not statistically significant}, overall the judges preferred \model over GPT-2 FT. 

\ARR{\subsection{Ablation Study}}
\ARR{
Here, we perform ablation studies to investigate the contribution of each component of \model.}

\noindent\textbf{Story Continuer}. \space
\ARR{To investigate the contribution of relationship LM and null LM, we construct an ablated baseline \model-$SLM$ where we use a single language model as the Story Continuer. This LM is trained to generate a sentence given a relationship or $\emptyset$. Table \ref{tab:ablate2} presents automatic evaluation of content quality and relationship faithfulness. We observe that conflating relationship LM and null LM into a single model leads to performance degradation in both content quality and relationship faithfulness. With \model-$SLM$, relationship faithfulness becomes more challenging to obtain with about 11\% lower \%Exact, 14\% higher \%Incorrect and 12\% lower P-CLS scores. Furthermore, In \model, Null LM has more freedom than Relationship LM in expressing content quality without any relationship constraints. Lack of this LM causes content quality to suffer in \model-$SLM$. This shows that both the LMs are necessary for the task of relationship-driven story generation.}
%

\begin{figure}
    \centering
    \includegraphics[width=0.75\columnwidth]{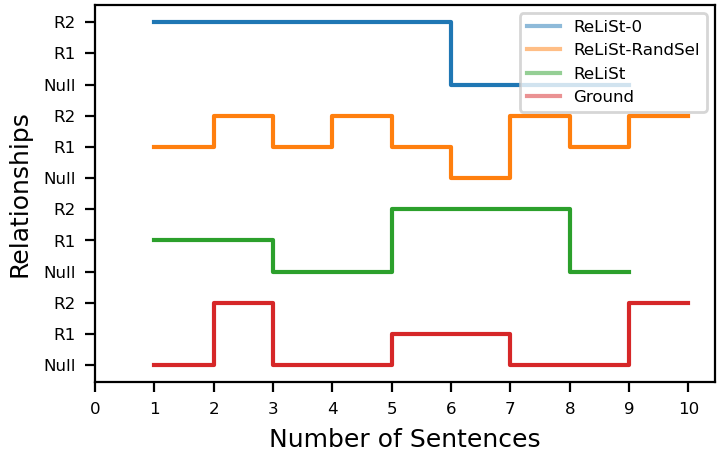}
    \caption{\ARR{An example of relationship flow across different models. R1 and R2 are two input relationships.}}
    \label{fig:flow}
\end{figure}

\noindent\textbf{Relationship Selector.} \space
To investigate the contribution of the relationship selector, we replace it with a random selector that randomly selects a relationship from $R$ (including $\emptyset$). We call this baseline \model-$RandSelect$.  
From Table \ref{tab:ablate2}, we can observe that randomly picking relationships hurts both content quality and relationship faithfulness. Figure \ref{fig:flow} plots the change in relationships ($R\in$\{R1, R2, $\emptyset$\}) throughout the story with different relationship selectors. \model-$RandSelect$ exhibits sharp and unnatural turns in relationships. 
This unnatural relationship flow impacts the story coherence negatively. \newline

\subsection{Analysis} 
In this section, we conduct analyses to gain insights into the working of \model. \\
\noindent\textbf{Case Study.} \space
Table \ref{tab:examples} shows two stories generated by \model and GPT-2 FT for the same input relationship set and prompt sentence. The input requires the story to express a positive relationships between ``Jonny'' and ``Curtis''. However, GPT-2 FT does not make any mention of ``Curtis'' and instead introduces ``Jonny'''s relationship with ``Mary''. \model continues the prompt sentence coherently while seamlessly introducing the relationship between ``Jonny'' and ``Curtis'' with accurate polarity (``love'' implying positive relationship). 
More examples of generated stories can be found in the Appendix Tables~\ref{tab:examples2}, \ref{tab:examples3}, and \ref{tab:examples7}.

While generating stories, \model assigns a latent relationship to each generated sentence. Next, we analyze the latent variables assignments to get insights into the generation process of the model.

\noindent\textbf{N-grams of each polarities.} \space 
We use the latent relationship assignments to analyze the word choices that \model makes to exhibit different relationships. For this, we consider the story sentences that had their latent relationship variable assigned as \textit{positive}, \textit{neutral} and \textit{negative}, and analyze the commonly generated $n$-grams ($n=\{1,2,3\}$). We find that \model uses $n$-grams like ``love'', ``help'',  and ``childhood friend'' in sentences in which it exhibits positive relationships. Similarly, it uses ``sees'' and ``meanwhile''; and ``kill'' and ``death'' in sentences in which it exhibits neutral and negative relationships respectively. This shows that using the latent variables, \model can  generate sentences that effectively express different relationships. Table \ref{tab:ngrams} in Appendix shows more $n$-grams.

\begin{figure}
    \centering
    \includegraphics[width=0.68\columnwidth]{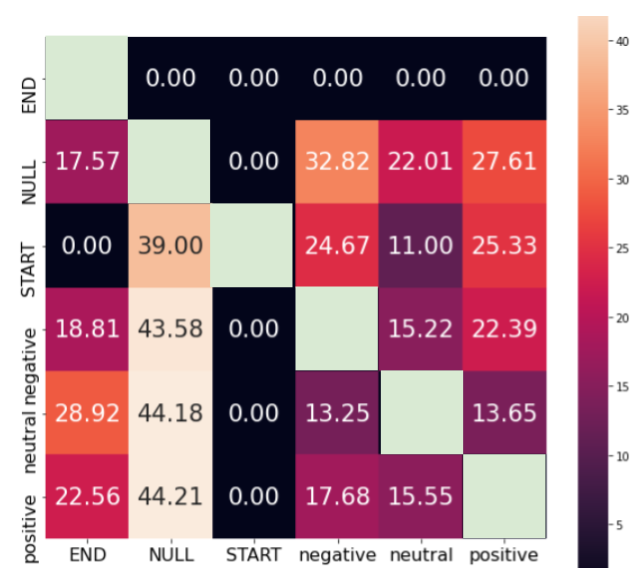}
    \caption{Transition probabilities illustrating changes in latent relationships within a story. All relationship polarities are most likely to be changed to $\emptyset$. Self-transitions are discounted to analyze relationship change.}
    \label{fig:transition}
\end{figure}

\noindent\textbf{Polarity analysis.} 
We use the latent variables to investigate any patterns in \model's tendency to exhibit different relationship polarities. First, we consider examples from the test set which have exactly three characters. Then, for each example, we create six different relationship sets using the three characters by assigning all possible combinations of \textit{positive}, \textit{neutral} and \textit{negative} relationships between them. We pair each of these relationship sets with the prompt for the test example and provide this input to \model to generate stories. Finally, We analyze the sequence of latent variable assignments from these generated stories. In Figure \ref{fig:transition}, we provide a heat-map for \model's probability to transition from the polarity on the $X$-axis to the polarity on the $Y$-axis. Transitions from a  polarity to itself are discounted.

We observe that after relationship selector chooses one of  \textit{positive}, \textit{neutral} and \textit{negative}, it assigns the next sentence to $\emptyset$ most frequently than any other polarity (discounting self-transitions). This could be because $\emptyset$ sentences are imperative for a smooth transition between sentences that describe input relationships. Details regarding this analysis are presented in Appendix.

\begin{table}
\centering
\footnotesize
\setlength\tabcolsep{12pt}
\renewcommand{\arraystretch}{1.1}
\begin{tabular}{lc}
\hline
Positions & \textbf{Pos / Neu / Neg / $\emptyset$}\\
\hline 
Beginning & $25.3\:/\:11.0\:/\:24.7/\:39.0$ \\
Ending & $24.7\:/\:24.0\:/\:21.0/\:30.3$ \\
Overall & $21.0\:/\:15.6\:/\:20.9/\:42.5$ \\
\hline
\end{tabular}
\caption{Distribution of relationship polarities in the beginning and ending sentences, and the overall story. \model's stories are more likely to start with positive or negative than neutral relationships but they are equally likely to end in any of the three polarities. }
\label{tab:6way}
\end{table}

Table \ref{tab:6way} presents polarity distributions for the first and the last generated sentences, as well as the overall story. Not all relationships are equally likely to be picked by the relationship selector of \model. Stories generated by \model are likely to continue from the prompt sentence with \textit{positive} ($25.3\%$) or \textit{negative} ($24.7\%$). Also, despite \model's overall propensity to avoid \textit{neutral} relationships (only $15.6\%$), it is as likely to end a story with this polarity ($24.0\%$) than with \textit{positive} ($24.7\%$) or \textit{negative} ($21.0\%$). 

Overall, the latent relationships of \model allows us to analyze relationship flows across the story, understand the vocabulary used to express different polarities and analyze the transition between relationships. %

\section{Conclusion}



This paper presents a new perspective in story generation-- from relationships. The proposed approach, \model, introduces relationships as latent variables for relationship-driven story generation. We jointly train the components of \model while addressing the major challenges of this task. \model outperforms baselines in generating coherent stories with desired relationships. Finally, we also observe how the latent variable based design of \model offers interpretability to the generation process without compromising its performance.
  

\section{Acknowledgement}
We thank the anonymous reviewers for their constructive feedback. This work was supported in part by NSF grant IIS-2047232.
\section*{Limitations}
\ARR{This work makes several assumptions about inter-character relationships that potentially hurt its expressivity. For example the work assumes that relationships are undirected, static, and can be expressed through polarity. Another limitation is that it also assumes that a story sentence can only express one relationship at a time. this could be an interesting future direction to explore. Also, note that our annotation pipeline for obtaining silver-labels uses a sentiment classification model. These models generally lose performance as the number of labels increases which in turn affects the quality of (silver) annotation and the generation. Hence, increasing the number of latent variables or granularity of relationship polarity is challenging.}

\ARR{We hope that future work can address these issues and our work can be a starting point in this exciting direction.} 

\section*{Ethical Considerations}
We train our model on a publicly available movie summary dataset that might contain (potentially harmful) social biases. Since, we have not employed any bias removal methods, model might replicate any biases found in the training data such as generating setting of the story based on the names of characters provided as inputs. Models' generated stories might also contain violent and graphic content, especially but not exclusively corresponding to ``negative'' relationships. The dataset is only in one language-- English.  

We conduct human evaluations on Amazon Mechanical Turk. For fairly compensating the workers for their efforts, the authors did several rounds themselves to calculate average time to finish one HIT. Based on the HIT timings, the workers were paid  \$11/hr. No personal, sensitive or identifying information was collected.

\bibliography{anthology,custom}
\bibliographystyle{acl_natbib}

\appendix

\section{Appendix}

 \subsection{Toolkits}
 We use NLTK toolkit Link: \url{https://www.nltk.org/} for computing BLEU scores and sentiment intensity. NLTK version is 3.6.2. Positive or Negative polarity is decided based on which of the intensity score is higher. Neutral score is decided if positive and negative sentiments have equal intensity. For ROUGE, we use \url{https://pypi.org/project/rouge/}. The version is 1.0.1. The f-measure score is used in ROUGE-1 and ROUGE-f. We use BookNLP toolkit \url{https://github.com/dbamman/book-nlp} for annotating our stories with charachter mentions and dependency parsing labels.

All results are based on a single batch of generated stories. P-CLS is an accuracy score averaged over three runs.

\subsection{Training and Inference Details}
\label{app:details}

\textbf{ Number of parameters}: GPT-2-medium has 345 million parameters. BERT-base has 110 million parameters. BART-large has 406 million parameters.

\textbf{GPU Details}: We use a NVIDIA GeForce RTX 2080 Ti machine to train and infer all our models.

\textbf{Polarity Classification (P-CLS)}: For training the polarity classifier, we finetune BERT-base~\cite{devlin-etal-2019-bert} on stories generated by the model. It takes as input \texttt{``Char1 and Char2 [SEP] $S$''} where $S$ is the generated story, and identifies the polarity of the relationship between characters \texttt{Char1} and \texttt{Char2}. The character pairs and the corresponding relationship polarity labels for training are obtained from the input relationships used to generate the story $S$. 

\subsection{Quality of \textit{silver labelled} dataset}
\label{app:quality-data}
\ARR{To assess the quality of the \textit{silver labelled} dataset, we compare our annotations to a subset of $105$ stories from the CMU Movie corpus \cite{srivastava2016inferring} that are manually annotated with inter-character relationships. We find that on an average $33.87 \%$ of automatically identified relationships are new. Among the remaining relationships which have matched character pairs with human-annotated data, $59.25 \%$ have the same and $40.74 \%$ have different polarity compared to human annotations. This shows that \textit{silver labelled} data has reasonable quality to be used for initizaliation.We also experimented with a BERT-based sentiment classifier, trained on SST to identify polarity of the relationships in place of the Sentiment Intensity Analyzer toolkit. However, the toolkit gave us the best match with human-annotated data. }

\subsection{Additional Human Evaluation Details}\label{app:human-eval}
\begin{figure*}
    \centering
    \includegraphics[width=\textwidth]{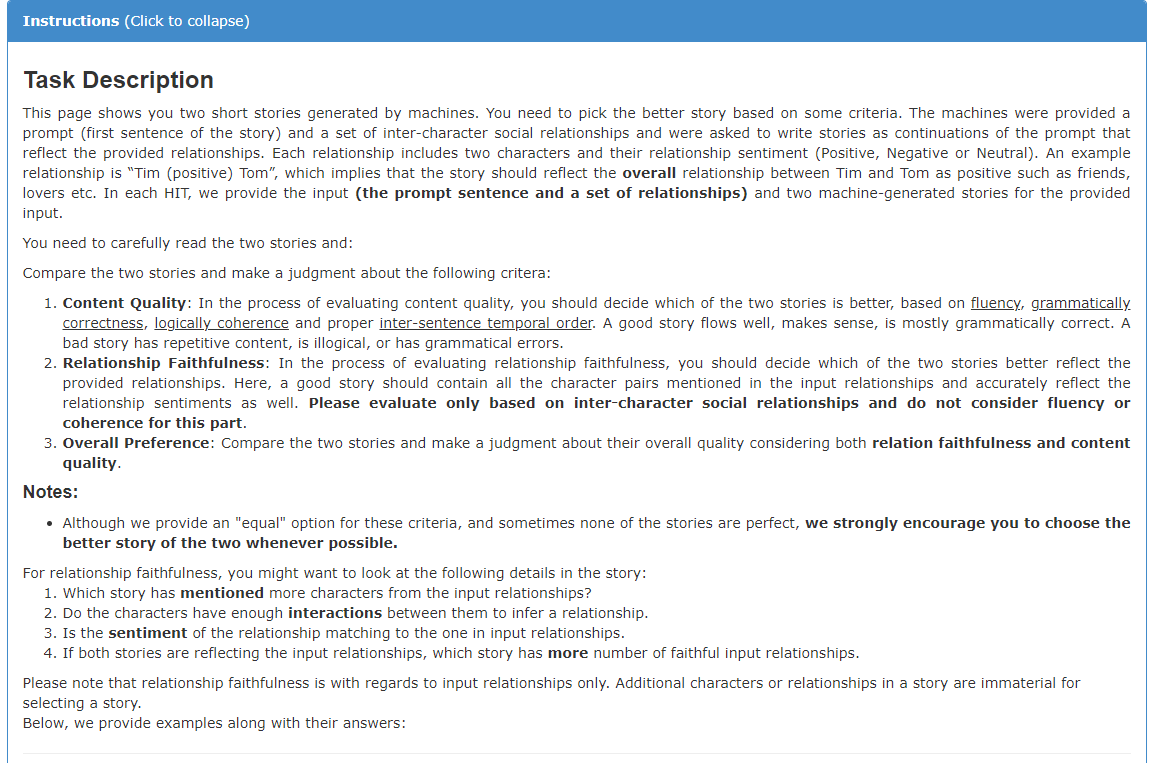}
    \caption{An screenshot of instructions for human evaluation on AMT}
    \label{fig:amt1}
\end{figure*}

\begin{figure*}
    \centering
    \includegraphics[width=\textwidth]{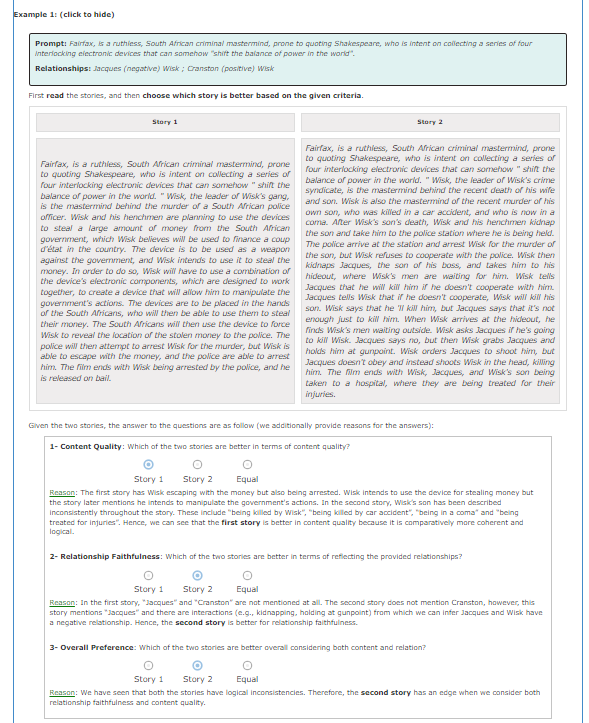}
    \caption{Continuing instructions for human evaluation on AMT}
    \label{fig:amt2}
\end{figure*}

Figures  \ref{fig:amt1} and \ref{fig:amt2} show the full set of instructions given to the participants. We filtered workers with those from US, UK or Canada and each of them should have done at least 5000 HITs. We have neither asked nor are aware of any other demographic information regarding them.
 
\subsection{Dataset License Details}
The dataset we have used was released under a Creative Commons Attribution-ShareAlike License.\footnote{http://www.cs.cmu.edu/~ark/personas/} Our research is consistent with the intended use. Proposed model trained on this dataset is for research use only, not commercial.

\subsection{Additional generation examples}
Table \ref{tab:examples2} shows example stories generated by \model and GPT-2 FT for the same input relationship set and prompt sentence. In the first example, GPT-2 FT makes no mentions of ``Jacques'' or ``Cranston''. In the story generated by \model, story continuer followed the prompt sentence naturally untill relationship selector chose ``Jacques'' and ``Wisk'''s relationship, which was manifested by story continuer in a negative interaction (``kidnapping''). Thereafter it continued the plot maintaining coherence and ``negative'' polarity between ``Jacques'' and ``Wisk''.  Tables \ref{tab:examples3} and \ref{tab:examples7} show more examples of GPT-2 FT and \model's generated stories.

\begin{table*}[!t]
\scriptsize
\renewcommand{\arraystretch}{1.2}
\begin{tabularx}{\textwidth}{lX}
     \hline
\textbf{\textcolor{forestgreen}{<positive>}} & love, help, friend, falls, friends, becomes, party, loves, marry, agrees, like, well, soon, still, fall love, best friend, old friend, united states, tries convince, agrees help, take care, falling love, true identity, still loves, childhood friend, comes back, fell love, confesses love, begins fall love 
\\ \hline
\textbf{<neutral>} & sees, meanwhile, gives, arrives, son, wants, years, room, daughter, relationship, meet, gives birth, manages get, wants marry, new life, goes meet, final scene, receives call, months later, gives birth son, two years later , home one day, three months later, decides get married 
\\ \hline 
\textbf{\textcolor{flame}{<negative>}} & kill, kills, killed, death, police, leave, tom, killing, fight, attempts, dead, gun, way 
tries kill, attempts kill, dead body, commits suicide, car accident, police officer, take revenge, one last, hotel room, becomes obsessed, decides leave, commit suicide, world war ii, decides take revenge 
\\ \hline 
\end{tabularx}
  \caption{Words and phrases that show how \model effectively exhibits different relationship polarities.}
  \label{tab:ngrams}
\end{table*}


\begin{table*}[h]
\footnotesize
\renewcommand{\arraystretch}{1.2}
\begin{tabularx}{\textwidth}{lX}
    \hline
    \textbf{$R$} & Jacques \textcolor{flame}{<negative>} Wisk ; Cranston \textcolor{forestgreen}{<positive>} Wisk \\
   \textbf{Prompt} & Fairfax, is a ruthless, South African criminal mastermind, prone to quoting Shakespeare, who is intent on collecting a series of four interlocking electronic devices that can somehow " shift the balance of power in the world. .\\
     \hline
\textbf{GPT-2 FT} &  Fairfax, is a ruthless, South African criminal mastermind, prone to quoting Shakespeare, who is intent on collecting a series of four interlocking electronic devices that can somehow " shift the balance of power in the world.  " Wisk, the leader of Wisk's gang, is the mastermind behind the murder of a South African police officer. Wisk and his henchmen are planning to use the devices to steal a large amount of money from the South African government, which Wisk believes will be used to finance a coup d'etat in the country. The device is to be used as a weapon against the government, and Wisk intends to use it to steal the money. In order to do so, Wisk will have to use a combination of the device's electronic components, which are designed to work together, to create a device that will allow him to manipulate the government's actions. The devices are to be placed in the hands of the South Africans, who will then be able to use them to steal their money. The South Africans will then use the device to force Wisk to reveal the location of the stolen money to the police. The police will then attempt to arrest Wisk for the murder, but Wisk is able to escape with the money, and the police are able to arrest him. The film ends with Wisk being arrested by the police, and he is released on bail. \\ \hline
\textbf{\model} & Fairfax, is a ruthless, South African criminal mastermind, prone to quoting Shakespeare, who is intent on collecting a series of four interlocking electronic devices that can somehow " shift the balance of power in the world. " Wisk, the leader of Wisk's crime syndicate, is the mastermind behind the recent death of his wife and son. Wisk is also the mastermind of the recent murder of his own son, who was killed in a car accident, and who is now in a coma. After Wisk's son's death, Wisk and his henchmen kidnap the son and take him to the police station where he is being held. The police arrive at the station and arrest Wisk for the murder of the son, but Wisk refuses to cooperate with the police. Wisk then kidnaps Jacques, the son of his boss, and takes him to his hideout, where Wisk's men are waiting for him. Wisk tells Jacques that he will kill him if he doesn't cooperate with him. Jacques tells Wisk that if he doesn't cooperate, Wisk will kill his son. Wisk says that he 'll kill him, but Jacques says that it's not enough just to kill him. When Wisk arrives at the hideout, he finds Wisk's men waiting outside. Wisk asks Jacques if he's going to kill Wisk. Jacques says no, but then Wisk grabs Jacques and holds him at gunpoint. Wisk orders Jacques to shoot him, but Jacques doesn't obey and instead shoots Wisk in the head, killing him. The film ends with Wisk, Jacques, and Wisk's son being taken to a hospital, where they are being treated for their injuries.\\
\hline
\hline
\textbf{$R$} & Henry \textcolor{forestgreen}{<positive>} Wayne \\
\textbf{Prompt} & Legendary bank robber Henry Manning pushes his luck too far and ends up in prison, where he suffers a massive stroke. \\
     \hline
\textbf{GPT-2 FT} & Legendary bank robber Henry Manning pushes his luck too far and ends up in prison, where he suffers a massive stroke.  He is sent to a mental institution, where he is placed in solitary confinement. He is given the choice of either serving time or serving time in prison. He chooses the latter, and is given a chance to live his life as a free man. After serving his sentence, Manning is released from solitary confinement and returns to New York City to begin his life anew. He meets a young woman named Jane, and they begin to fall in love. However, Manning's life is in danger, and he is forced to choose between his love for Jane and his life in prison, or he will be sent to prison for life. The film ends with Manning being released from prison and returning to his old life.\\ \hline
\textbf{\model} & Legendary bank robber Henry Manning pushes his luck too far and ends up in prison, where he suffers a massive stroke. Wayne, who has been in prison for 20 years, decides to help him by giving him a chance to redeem himself for his crime.
 \\
    \hline
     \hline
\end{tabularx}
  \caption{Two examples of generated stories from \model and GPT-2 FT. The examples encapsulates strengths and weakness of the models. In first example neither ``Jacques'' or ``Cranston'' is present in GPT-2 FT generated story. \model missed ``Cranston'' but captured Jacques's relationship to Wisk via ``kidnapping''. In second example, GPT-2 FT generated a new character ``Jane'' and a new relationship over faithfully describing the input relationship like \model.}
  \label{tab:examples2}
\end{table*}

\begin{table*}[h]
\footnotesize
\renewcommand{\arraystretch}{1.2}
\begin{tabularx}{\textwidth}{lX}

    \hline
\textbf{$R$} &  Amarilly \textcolor{flame}{<negative>} Terry ; Mrs. \textcolor{forestgreen}{<positive>} Terry ; Amarilly \textcolor{forestgreen}{<positive>} Mrs. \\
\textbf{Prompt} & Set in San Francisco during the early 1900s , the film revolves around Amarilly , the daughter of a widowed scrubwoman .   \\
     \hline
    \textbf{GPT-2 FT} &    Set in San Francisco during the early 1900s, the film revolves around Amarilly, the daughter of a widowed scrubwoman.  Amarilly's mother, Mrs. Davenport, is a widower who lives with her mother's two children. Amarilly and her mother live in a dilapidated bungalow on the outskirts of San Francisco, where Amarilly works as a housekeeper. Amarillie's father is a wealthy businessman, and Amarilly has a son named Terry, who is the son of a wealthy widow. Terry's parents are both widows, and Terry is the only child of the wealthy widow's husband, who has a daughter named Amarilly. The film begins with the arrival of Amarilly at a boarding school, where she is introduced to the school's principal, Mr. Pritchett, and the school principal's daughter, Ms. Parnell. The principal, however, is not impressed with Amarilly, and tells her that she is not welcome in the school, and that she will be expelled from the school if she does not comply with the school rules. She also tells the principal that she doesn't want to be a part of the school and will not be allowed to participate in any activities that would interfere with her academic progress. The next day, the principal tells Amarilly that she has to leave the school immediately, and asks her to stay at the boarding school for a few days. When Amarilly arrives, she is greeted by the principal, who tells her to leave immediately. She does so, and is told that the school will be closed for the day, and she will have to return to school the next day. When she returns to school, she finds that the principal has told her that Amarilly is not allowed to enter the school. She goes to the principal and tells him that she wants to stay, but the principal does not believe her, and says that he will not allow her to attend the school unless she agrees to go with him. She leaves the school the following day, but when she returns, she discovers that her father has left her. She tells her father that he is going to take her back to San Francisco to live with him, but he refuses to let her go, saying that he does not want to see her again. She then goes to her father's house, where he tells her he is leaving for San Francisco and that he wants her to come with him to stay with him for the rest of her life. The following morning, Amarilly returns to her parents'house, and they tell her that they will not let her stay with them for the remainder of their lives. When they return to the house, they find that the house has been ransacked and that the only way to get out of the house is to find a way out of it. They go to the attic, where they find a large amount of money in the attic. They find the money, and discover that the money has been stolen from the house. They take the money to the\\ \hline
\textbf{\model} & Set in San Francisco during the early 1900s, the film revolves around Amarilly, the daughter of a widowed scrubwoman. She lives with her widowed mother and younger brother. Amarilly's mother is a widower, and Amarilly is the only child of her mother's deceased husband, who died in childbirth. When Amarilly's mother dies, Amarilly and her younger brother, Terry, move in with their mother's widowed father, who lives in the same house as her mother. Amarilly grows up to be a good - natured, kind - hearted young man, but her father is a cruel and cruel man, and he has no respect for Amarilly. The film begins with Amarilly's mother's death, and the film then moves to the present day, where Amarilly finds out that she is pregnant with Terry's child. Amarilly decides to take the baby with her, and she and Terry decide to have an abortion. After the abortion, Amarilly's mother dies, and she decides to go back to her mother's house and live with her father. 
 \\
    \hline
    
    \hline
\textbf{$R$} & Daya \textcolor{flame}{<negative>} Loganathan ; Daya \textcolor{flame}{<negative>} Satya ; Kiruba \textcolor{flame}{<negative>} Satya \\
\textbf{Prompt} &  Satya Narain and Kiruba are close friends living opposite each other in the neighbourhood.  \\
     \hline
    \textbf{GPT-2 FT} &   Satya Narain and Kiruba are close friends living opposite each other in the neighbourhood.  Satya's father is a successful businessman. Kiruba's mother is a social worker. Satya and Satya fall in love, but Satya can't bear the thought of marrying Kiruba. He wants to marry Kiruba, but Kiruba refuses to marry Satya. When Satya asks Kiruba to marry him, Kiruba says that he doesn't want to marry anyone else, and that he's not interested in marrying anyone else. When Kiruba asks Satya to marry her, Satya tells him that he won't marry anybody else. He tells Satya that if he does, he 'll marry her. He then tells Kiruba that he wants to be married to Satya, but he ca nâ€™t marry her because of his love for her. After Satya leaves, he goes to Kiruba â€™s house and tells her that he loves her, and asks her to come to his house. She refuses, and he tells her to stay with him. He goes to her house, and she tells him she wants to go with him, and tells him to come with her. She tells him not to go, and says that she wants him to be with her, so that he can get married to her. The film ends with Satya getting married to a beautiful young woman named Kiruba in the middle of the night.\\ \hline
\textbf{\model} & Satya Narain and Kiruba are close friends living opposite each other in the neighbourhood. One day, Satya and Daya get into a fight and Satya beats him up badly. Daya's father comes to Satya's house and tells him that he is going to take revenge on his son for what he has done. Satya goes to the police station and tells the truth to Daya. Daya's father beats up Satya's mother and he is arrested by the police. The next day, Satya's mother comes to the hospital and tells Satya that she is pregnant. Satya tells her that he doesn't want to see her again and that he wants to marry Kiruba. She tells him to go back to his house and tell his mother that he loves her. Satya then goes to Kiruba's house and asks him if he loves Kiruba. Kiruba replies that he doesn't love her. Satya's mother then tells Kiruba that she loves him and that she wants him to marry her. When Satya asks Kiruba why he doesn't love Kiruba, Kiruba says that he does not love her. Satya's mother then says that Kiruba does love her and asks her to marry him. At the end of the movie, Daya and his father come to meet Kiruba and ask him to come and live with them. 
 \\
    \hline
     \hline
\end{tabularx}
  \caption{Additional Examples}
  \label{tab:examples3}
\end{table*}

\begin{table*}[h]
\footnotesize
\renewcommand{\arraystretch}{1.2}
\begin{tabularx}{\textwidth}{lX}
    \hline
    \hline
\textbf{$R$} & Dan \textcolor{flame}{<negative>} Tucker, Dan \textcolor{flame}{<negative>} Max \\
\textbf{Prompt} & The film 's plot is loosely adapted from " The Austin Road Trip Story " in Max 's book I Hope They Serve Beer in Hell. \\
     \hline
    \textbf{GPT-2 FT} &  The film's plot is loosely adapted from " The Austin Road Trip Story " in Max's book I Hope They Serve Beer in Hell.  The film begins with a young man named Max, who has just graduated from high school and is about to start a new life with his family. He meets Max's older brother, Tucker, at a bar, and they begin to fall in love. However, Tucker's father, who is also a drug addict, has a change of heart and decides to take Max to a rehab center. Max and Tucker go to the rehab center, where they meet a group of drug - addicted people, including a drug - addict named Maxine, who also happens to be a drug dealer. The drug dealer, Maxine's ex-boyfriend, is also involved in the drug deal, and Max is forced to deal with the drug dealer's gang. Max, Tucker and Maxine get into a fight, but Max manages to escape. Max then goes to the hospital, where he is treated for his injuries. The next day, Max meets up with Maxine at the hospital. He tells her that he is going to take her to rehab, and she agrees to go with him. The film then cuts to the beginning of the film, where Max is in the hospital recovering from his injuries, and he tells Max that he wants to go to rehab. Max tells him that he doesn't want to go, and that he's going to do whatever it takes to get there. The movie then cuts back to the end of the movie, where Tucker is in a hospital bed, with Max in his arms. Max asks him if he 'd like to go back to school, and Tucker replies, " No, I'm not going to go. " Max then asks Max if he would like to see him again. Max says yes, and the film ends with the two of them walking down the street.\\ \hline
\textbf{\model} & Dan Tucker, a high school senior, is on his way to a party with his friends when he is accosted by a group of bullies. The bullies drag him to the side of the road and beat him savagely with a tire iron. Dan tries to get help, but is beaten up by the bullies. The next day, Dan and his friends go to a bar for a drink, where they meet up with Tucker. Tucker tells Dan that he's going to kill himself, but Dan is reluctant to do so, as he feels that killing himself would be too painful. The next morning, Dan goes to the hospital to see Max, who is recovering from his injuries. Dan tells Max that he's going to commit suicide, but Max says that he can't, because he 'd be too sick to do it. Dan then goes to Max's house, and tells him that he has to go kill himself, because his parents won't let him. Then, Dan's parents show up, and they tell him that they're willing to let him go, because they know that he's going to die soon. Then, they show up at his house, and he tells them that he doesn't want them to kill him, because that's the only thing that he knows how to do, and that they won't be able to stop him from killing himself. He then tells them he's going to take them to his house, and that he 'll kill himself if they don't. They then go to Max's house, where they find that his parents are there. They tell them that they're not going to let them kill him, because that's the only way they can stop him. They then leave, and Dan and Max go back to the house, where they see that Max's parents are gone. They go to Dan's house, where Max's parents are waiting for them. Max tells them to go to his room, and they do, and then they go to the bathroom, where Dan is waiting for Max. He says that they're going to have to kill themselves, because they've never done it before, and they 'll never do it again. He then says that if they don't, then he's going to be the one to kill them, and that's what he wants them to do. The film ends with Dan's parents talking about how much they love him, and how much he's going to love them, and that's what they want him to do.
 \\
    \hline
    \hline
\textbf{$R$} & Katia \textcolor{flame}{<negative>} Mikos\\
\textbf{Prompt} &  The plot involves a near invincible man, known only as Mikos, a dead Greek serial murderer, whose blood coagulates incredibly fast, but this gift has made him insanely homicidal.   \\
     \hline
    \textbf{GPT-2 FT} &    The plot involves a near invincible man, known only as Mikos, a dead Greek serial murderer, whose blood coagulates incredibly fast, but this gift has made him insanely homicidal.  Mikos is the son of a wealthy Greek family, and his father is the head of the Greek mafia. Mikos'father is a wealthy businessman, and Mikos lives with his father's wife and two young daughters. When Mikos discovers that his father has been murdered, he decides to take revenge on his father. He kidnaps Mikos and takes him to the Greek underworld, where he is forced to fight his way out of the clutches of the mobsters. The film opens with Mikos fighting his way through the streets of the underworld. He is eventually captured by the police, and is taken to the police station, where Mikos confesses that he killed his father, and that he was the one who killed his mother. The police then arrest Mikos, and he is sentenced to life in prison. The next day, Mikos visits his father in prison, where his father tells him that he is going to kill him, but Mikos refuses, saying that he will never kill him. The scene then cuts to Mikos walking down the street, and the scene cuts back to the scene of the murder of his father and his wife.\\ \hline
\textbf{\model} & The plot involves a near invincible man, known only as Mikos, a dead Greek serial murderer, whose blood coagulates incredibly fast, but this gift has made him insanely homicidal. Mikos kills Katia, the daughter of a Greek politician, in a fit of rage. Katia's body is found, and Mikos is convicted of the murder. He is sentenced to 15 years in prison for the murder of Katia's father. Mikos's sentence is that he will never see Katia again, and he will be sent to a mental institution. In prison, Mikos meets Katia's mother, who tells him that Katia is dead, and that she is the one who killed her father. Mikos tells Katia that he is going to kill her, but Katia doesn't believe him. The next day, Mikos goes to Katia's mother's house, where he finds her dead body. Katia tells Mikos that she will never forgive him for killing her father, and that if he does, she will kill him too. The film ends with Mikos and Katia walking down the street together.
 \\
    \hline
     \hline
\end{tabularx}
  \caption{Additional Examples}
  \label{tab:examples7}
\end{table*}

\end{document}